\pdfoutput=1

\documentclass[11pt]{article}

\usepackage[preprint]{acl}

\usepackage{times}
\usepackage{latexsym}

\usepackage[T1]{fontenc}

\usepackage[utf8]{inputenc}

\usepackage{microtype}

\usepackage{inconsolata}

\usepackage{graphicx}

\usepackage{hyperref}       
\usepackage{url}            
\usepackage{booktabs}       
\usepackage{amsfonts}       
\usepackage{nicefrac}       
\usepackage{microtype}      
\usepackage{xcolor}         

\usepackage{amsmath}
\usepackage{multirow}
\usepackage{array}
\usepackage{enumitem}
\usepackage{booktabs}
\usepackage{svg}
\usepackage{caption}
\usepackage{graphicx}
\usepackage{wrapfig}
\usepackage{float}
\usepackage{subcaption}
\usepackage{hyperref}
\usepackage{xcolor}
\usepackage{soul}

\usepackage{algorithm}
\usepackage{algpseudocode}

\title{InsBank: Evolving Instruction Subset for Ongoing Alignment}

\author{
 \textbf{Jiayi Shi\textsuperscript{1}}\footnotemark[2], \hspace{0.4cm}
 \textbf{Yiwei Li\textsuperscript{1}}\footnotemark[2], \hspace{0.4cm}
 \textbf{Shaoxiong Feng\textsuperscript{2}}, \hspace{0.4cm}
 \textbf{Peiwen Yuan\textsuperscript{1}}, \hspace{0.4cm}
 \textbf{Xinglin Wang\textsuperscript{1}}, \hspace{0.4cm}
\\
 \textbf{Yueqi Zhang\textsuperscript{1}}, \hspace{0.4cm}
 \textbf{Chuyi Tan\textsuperscript{1}}, \hspace{0.4cm}
 \textbf{Boyuan Pan\textsuperscript{2}}\footnotemark[3], \hspace{0.4cm}
 \textbf{Huan Ren\textsuperscript{2}}, \hspace{0.4cm}
 \textbf{Yao Hu \textsuperscript{2}}, \hspace{0.4cm}
 \textbf{Kan Li\textsuperscript{1}}\footnotemark[3], \hspace{0.4cm}
\\
\\
 \textsuperscript{1} School of Computer Science, Beijing Institute of Technology \\
 \textsuperscript{2} Xiaohongshu Inc
\\
 \texttt{\{shijiayi,liyiwei,peiwenyuan,wangxinglin\}@bit.edu.cn} \\
    \texttt{\{shaoxiongfeng2023\}@gmail.com} \  \texttt{\{panboyuan,zengshu,xiahou\}@xiaohongshu.com} \\
     \texttt{\{zhangyq,tanchuyi,likan\}@bit.edu.cn} 
}

\begin{document}
\maketitle

\renewcommand{\thefootnote}{\fnsymbol{footnote}} 
\footnotetext[2]{Equal contributions.} 
\footnotetext[3]{Corresponding authors.} 

\renewcommand{\thefootnote}{\arabic{footnote}}

\begin{abstract}
Large language models (LLMs) typically undergo instruction tuning to enhance alignment. Recent studies emphasize that quality and diversity of instruction data are more crucial than quantity, highlighting the need to select diverse, high-quality subsets to reduce training costs. However, how to evolve these selected subsets alongside the development of new instruction data remains insufficiently explored. To achieve LLMs' ongoing alignment, we introduce Instruction Bank (\textbf{InsBank}), a continuously updated repository that integrates the latest valuable instruction data. We further propose Progressive Instruction Bank Evolution (\textbf{PIBE}), a novel framework designed to evolve InsBank effectively and efficiently over time. PIBE employs a gradual data selection strategy to maintain long-term efficiency, leveraging a representation-based diversity score to capture relationships between data points and retain historical information for comprehensive diversity evaluation. This also allows for flexible combination of diversity and quality scores during data selection and ranking. Extensive experiments demonstrate that PIBE significantly outperforms baselines in InsBank evolution and is able to extract budget-specific subsets, demonstrating its effectiveness and adaptability.\footnote{Our code has been released on \url{https://github.com/jiayinlp/InsBank}}

\end{abstract}
\section{Introduction}

Instruction fine-tuning is widely adopted to refine pre-trained LLMs to accurately understand human instructions and provide precise, pertinent and harmless responses \citep{collection-flan-2022, ds-survey}. LIMA \citep{ift-lima} has proved that the quality and diversity of instruction data are significantly more critical than its sheer quantity for training, motivating recent efforts in instruction data selection to reduce unnecessary training costs by eliminating low-quality and redundant data \citep{ds-survey}. However, how to evolve the selected instruction subset in parallel with the development of the instruction data remains underexplored.

\begin{figure}[t]
\begin{center}
\includegraphics[width=0.98\columnwidth]{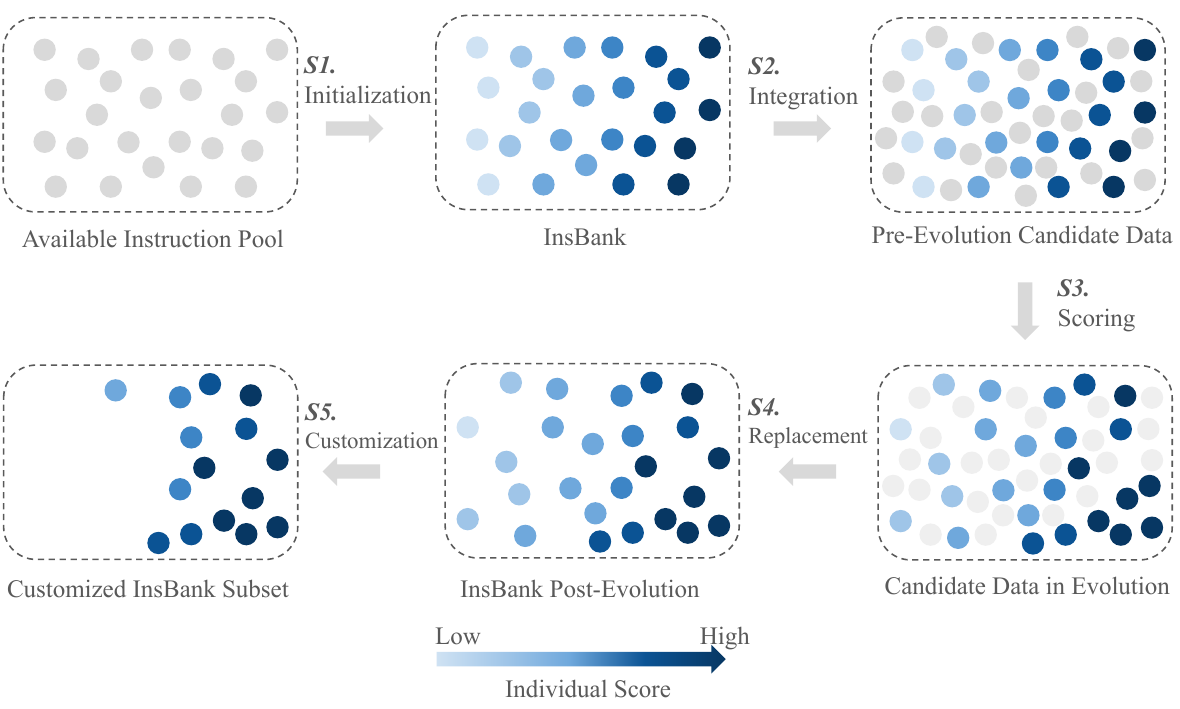}
\end{center}
\caption{Illustration of InsBank evolution. It is initialized by data selection on all current available instruction data, and it will evolve itself as long as new instruction data are proposed. A smaller training subset can be obtained from InsBank according to user training budget.} 
\label{fig: data selection}
\end{figure}

Specifically, with the continuous emergence of instruction datasets (The timeline of part instruction datasets is shown in Appendix~\ref{appendix: timeline}), it becomes necessary to regularly update the instruction subset to incorporate the latest advanced instruction data in order to ensure ongoing improvements in the alignment capabilities of LLMs. Simultaneously, the subset size must be controlled to avoid excessive growth that could lead to increased training costs. To address these practical challenges, we propose a novel concept termed \textbf{InsBank} (Instruction Bank). 
InsBank is designed to support instruction subset evolution with two key properties: 
(1) To prevent unbounded growth, InsBank maintains a constant size by replacing low-quality old samples with an equal number of high-scoring new ones during evolution.
(2) Samples in InsBank are ranked according to their overall scores to enable users to extract subsets that are tailored to specific training budgets, simply by selecting the top-ranked samples. 
The evolution process of InsBank is illustrated in Figure~\ref{fig: data selection}.

As the scale of existing instruction sets continues to grow\citep{ds-survey, collection-flan-2022, collection-self-instruct, collection-wizardlm}—reaching millions or even billions of instances—the cost of exhaustively traversing all candidate data during each InsBank evolution becomes prohibitively high. To address this challenge, we propose Progressive Instruction Bank Evolution (\textbf{PIBE}), a method designed for continuous and efficient selection of the optimal instruction subset. PIBE evolves InsBank in a gradual manner, ensuring long-term efficiency. Unlike the naive approach, it significantly reduces the cost of evolution by excluding previously filtered-out data and focusing solely on newly proposed samples and the current InsBank. 

Additionally, the orderliness of InsBank calls for an overall score that integrates both individual quality and diversity signals. While quality scores can be readily obtained through manual or model-based annotation, measuring individual diversity requires global comparisons among candidates. Unfortunately, existing instruction data selection methods struggle to effectively represent and combine quality and diversity for ranking purposes. This challenge is further exacerbated by the absence of historical data, which alters the distribution of candidates and underscores the need to retain historical data distributional information during evolution. Existing diversity-driven data selection methods \citep{ds-deita, ds-self-evolve} typically fall into two categories: k-nearest neighbor (k-NN) approaches \citep{indiv_eval_semantic1} and geometry-based coreset sampling methods \citep{indiv_eval_coreset1}. Both of them rely exclusively on local information from a limited number of neighboring points, which limits their ability to capture global relationships and provide reliable individual diversity scores for ranking. Furthermore, they lack mechanisms to preserve information about previously discarded data, making them ill-suited for progressive selection.
Inspired by Affinity Propagation \citep{cluster-ap}, we frame InsBank data selection as an exemplar election process, where the representativeness of each data point is quantified through an iterative voting mechanism. The representativeness further serves as the individual diversity score, and the voting results are passed to the next iteration as historical information to preserve the distribution of absent data. Moreover, existing data selection methods either prioritize quality or diversity \citep{ds-alpagasus}, or address them sequentially \citep{ds-deita}, failing to consider both aspects equally. Conversely, our diversity score integrates seamlessly with the quality score, enabling comprehensive and flexible instruction selection and InsBank ranking.

We simulate the instruction set development with five datasets and perform InsBank evolution on them with PIBE and we elaborate on the rationale for selecting these datasets in Appendix~\ref{appendix: Justification of Data Composition}. We evaluate the general instruction following capability of fine-tuned models on AlpacaEval \citep{benchmark-alpaca-eval}, MT-Bench \citep{benchmark-mtbench}, IFEval \citep{benchmark-ifeval}, OpenLLM Leaderboard \citep{benchmark-open-llm-leaderboard-v1} and FollowBench \citep{benchmark-followbench}. Experimental results show that PIBE outperforms the baselines and successfully evolves the instruction bank in parallel with the development of instruction sets. Besides, analysis on orderliness of InsBank indicates that users can flexibly select a smaller subset based on their budget.
Ours contributions can be summarized as follows:
\begin{itemize}[nosep]
    \item We propose InsBank, a dynamic framework for evolving instruction subsets alongside the development of instruction data, enabling continuous alignment improvements.

    \item We develop Progressive Instruction Bank Evolution (PIBE), an efficient approach that leverages a memory-enhanced diversity score and seamlessly integrates it with quality scores for optimal subset selection.

    \item We introduce a unified scoring system for individual samples, ensuring an ordered InsBank and enabling flexible extraction of high-quality subsets tailored to user budgets.
    
    \item Extensive experiments demonstrate that PIBE not only outperforms baseline methods in evolving InsBank but also provides flexible, budget-aware data selection, highlighting its effectiveness and adaptability.
\end{itemize}

\section{Preliminaries}
\label{appendix: preliminary}
\subsection{Instruction Data Selection Problem}
Following \citet{ds-deita}, given a collection of instruction data $\mathcal{X} = \{x_1, x_2, ..., x_n\}$ where $x_i$ is an individual instruction-response pair, data selection selects an instruction subset $\mathcal{P}_{\pi}^m$ of size $m$ from $\mathcal{X}$, where $\pi$ is the data selection strategy. Denote the performance evaluation function for $\pi$ as $Q$, the optimal data selection strategy $\pi ^*$ with subset size $m$ satisfies:

\begin{equation}
\small
    \pi^* = \arg \max_{\pi}Q(\mathcal{P}_\pi^m)
\end{equation}

\subsection{Selection Metrics}
Previous research \citep{ds-deita, ds-survey} highlight that the effectiveness of instruction set selection depends on both quality and diversity. In line with this, we focus on the two aspects in this paper:

\textbf{Quality} of instruction data primarily refers to the accuracy and rationality which estimate the consistency and coherence of the instruction context, as well as whether the response accurately corresponds to the instructions \citep{ds-survey}. In this work, we adopt the quality evaluation model of DEITA \citep{ds-deita} for quality annotation. 

\textbf{Diversity} of instruction data is critical to the generalization ability of the trained model \citep{ds-survey}. There are currently two major approaches to measure diversity: k-nearest neighbor (k-NN) \citep{indiv_eval_semantic1} and geometry-based coreset sampling \citep{indiv_eval_coreset1}. 
The kNN approach measures sample's diversity by its distance to its $j$-th k-nearest neighbor (k-NN) with the help of text embeddings as shown in Eq.~\ref{eq: knn}:
\begin{equation}
\small
\label{eq: knn}
    kNN_i^j = d(e(x_i), e(N_j(x_i)))
\end{equation}
where $N_j(x_i)$ denotes the $j$-th closest neighbor of $x_i$ in the embedding space projected by $e(\cdot)$, and $d(\cdot, \cdot)$ calculates the distance between $x_i$ and $N_j(x_i)$.
The geometry-based coreset sampling approach is to find the most informative-and-diverse subset that represents the entire dataset the most through controlling the minimum distance between any two samples for subset selection \citep{indiv_eval_coreset1, indiv_eval_coreset2}. However, both methods rely solely on local information from nearby points, making it difficult to capture the global distribution relationships or utilize historically eliminated points, resulting in inadequate individual diversity scores for subset evaluation.

\subsection{Affinity Propagation}
\label{sec: ap}

Affinity Propagation (AP) \citep{cluster-ap} is a clustering algorithm that leverages message-passing to uncover the global distribution of data. It identifies exemplars by iteratively transmitting two kinds of messages between data points:

\begin{itemize}[nosep]
    \item \textbf{Responsibility ($R[i,k]$)} This message sent from point \(i\) to point \(k\) represents how suitable point \(k\) is to serve as the exemplar for point \(i\).
    \item \textbf{Availability ($A[i,k]$)} This message sent from point \(k\) to point \(i\) represents how appropriate it would be for point \(i\) to choose point \(k\) as its exemplar, taking into account the current responsibilities sent from other points to \(k\).
\end{itemize}

The messages are updated iteratively based on the rules as shown in Eq.~\ref{eq: ap}.
Here, \(S[i,k]\) represents the similarity between point \(i\) and point \(k\) where $i \neq k$. And $S[k,k]$ is filled by the predefined preference value which represents the preference for sample $i$ as an exemplar.

\begin{equation}
\small
\label{eq: ap}
    \begin{aligned}
        R[i, k] &\leftarrow S[i, k] - \max_{k' \neq k} \left\{A[i, k'] + S[i, k']\right\}, \\
        A[i, k] &\leftarrow \min \left\{0, R[k, k] + \sum_{i' \notin \{i, k\}} \max \left\{0, R[i', k]\right\}\right\}, \\
        A[k, k] &\leftarrow \sum_{i' \neq k} \max \{0, R[i', k]\}, \\
    \end{aligned}
\end{equation}

At any given moment, the clustering result can be determined by summing $R$ and $A$. For $x_i$, let $k'$ be the index that maximizes $A[i,k] + R[i,k]$, the conclusion are as follows: (1) if $i=k'$, then $x_i$ is a cluster center, (2) if $i \neq k'$, then $x_i$ belongs to the cluster center $x_{k'}$. That is, for $R+A$, the $i$-th row represents the votes cast by $x_i$ for different points to represent itself, while the $j$-th column represents the votes received by $x_j$. Based on this, we obtain the representativeness of $x_i$ according to the voting results by subtracting the votes cast by $x_i$ for other samples from the votes received by $x_i$. This result serves as individual diversity score.

\section{Progressive Instruction Bank Evolution}

\begin{figure*}[hbtp]
\begin{center}
\includegraphics[width=0.84\textwidth]{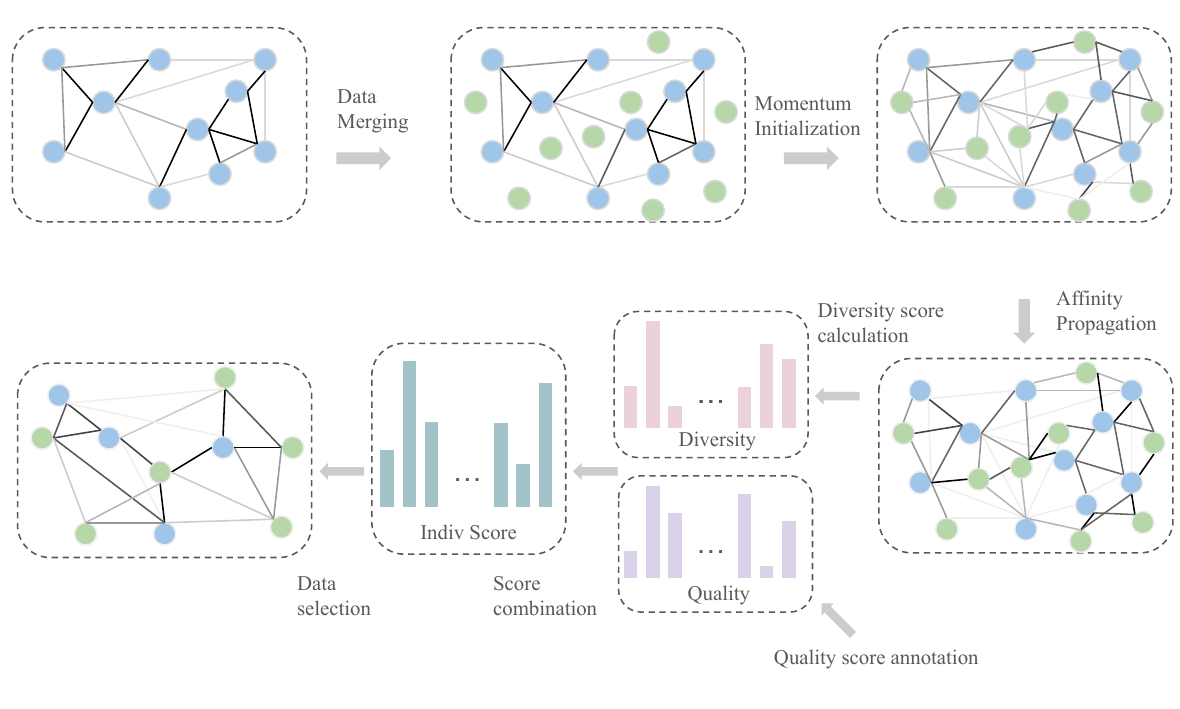}
\end{center}
\caption{The framework of PIBE begins by integrating newly proposed candidates with the existing InsBank data and initializing momentum information based on historical records. Then, affinity propagation incorporating the momentum is applied to compute diversity scores. Subsequently, the quality scores obtained via model-based annotation are combined with the diversity scores to produce an individual overall score. Finally, the top-$k$ samples with the highest overall scores are selected to form the evolved InsBank where $k$ is the budget.}
\label{fig: PIBE}
\end{figure*}

In this section, we provide a detailed explanation of PIBE, whose pipeline is depicted in Figure~\ref{fig: PIBE}.

\subsection{Gradual Evolution Formulation}

In this work, we propose the instruction subset evolution task to build the InsBank. Denoting current available instruction data as $\mathcal{X}_0$, the instruction bank $\mathcal{B}_{\pi}^{0,m}$ of size $m$ is initialized through data selection which can be presented as $\mathcal{B}_{\pi}^{0,m} = \pi(\mathcal{X}_0)$. Then, when new instruction dataset $\mathcal{X}_1$ is proposed, $\mathcal{B}_{\pi}^{0,m}$ should evolve itself to adapt to changes in data distribution. The naive manner of InsBank evolution can be represented as $\mathcal{B}_{\pi}^{1,m} = \pi(\mathcal{X}_0, \mathcal{X}_1)$ which can be extended to $\mathcal{B}_{\pi}^{t+1,m} = \pi(\mathcal{X}_0, ..., \mathcal{X}_t, \mathcal{X}_{t+1})$ for future evolution. However, this manner requires substantial storage and computational resources to calculate diversity scores as $t$ continues to increase. To improve the long-term evolution efficiency, we propose a gradual manner where only the newly proposed instruction data $\mathcal{X}_{t+1}$ along with the data participated in last round of evolution $\mathcal{X}_t + \mathcal{B}_{\pi}^{t-1,m}$ are involved into the current round of evolution, and the evolution can be represented as $\mathcal{B}_{\pi}^{t+1,m} = \pi(\mathcal{X}_{t+1}, \mathcal{X}_t + \mathcal{B}_{\pi}^{t-1,m})$.

In addition to the update of InsBank, we evaluate the diversity and quality of each sample $x_i$ and provide an overall individual score for data ranking. Users can quickly select a smaller subset according to the data ranking to suit their own training budget.

\subsection{Historical Information Flowing}

Although a large amount of data is eliminated during InsBank evolution for efficiency, preserving their distribution information is crucial for maintaining InsBank's global representativeness. To address this, we introduce a momentum matrix based on historical voting results to retain the distribution information of excluded data, which flows across iterations, allowing filtered-out data to re-engage in future exemplar selection and preventing suboptimal global representativeness.

As described in Section~\ref{sec: ap}, we evaluate individual diversity through AP. By analyzing the similarity between previously selected data and newly proposed candidates, we estimate the suitability of new data as exemplars for the existing data and vice versa, represented by the responsibility matrix. 

Formally, let $\mathcal{X}'_t = \mathcal{X}_t \cup \mathcal{B}^{t-1, m}_\pi$ denote the full candidate data set from the previous round of InsBank evolution, and $\mathcal{X}'_{t+1} = \mathcal{X}_{t+1} \cup \mathcal{B}^{t, m}_\pi$ denote the full candidate data set of the $(t+1)$-th evolution round. Then, the matrix $Sim_{t+1}$ of size \( |\mathcal{X}'_t| \times |\mathcal{X}_{t+1}| \) represents the cosine similarity between $\mathcal{X}'_t$ and $\mathcal{X}_{t+1}$.
Given the historical information matrix \(H_t\) of size \(|\mathcal{X}'_t| \times |\mathcal{X}'_t|\), representing the responsibility matrix stored from the \(t\)-th round of InsBank evolution, we derive the momentum responsibility matrix \(M_t\) using \(H_t\) and \(Sim_{t+1}\):
\begin{equation}
\label{eq: estimate top-right}
\small
\begin{aligned}
        w_{jk} &= \frac{Sim[j,k]}{\sum_{l=1}^{|X'_{t}|} Sim[l,k]}, \\
        M_t[i,k] &= \sum_{j=1}^{|X'_{t}|} w_{jk} * R_t[i,j]     
\end{aligned} 
\end{equation}

\begin{equation}
\small
\label{eq: estimate bottom-left}
        M_t[i,k] = \sum_{j=1}^{|X'_{t}|} w_{ij} * R_t[j,k]    
\end{equation}
This allows the filtered-out data to participate in exemplar election during future history-aware AP processes.

The structure of $M_t$ is depicted in Appendix~\ref{appendix: momentum-responsibility-matrix}. The top-left part of $M_t$ contains responsibility values between data in \(\mathcal{B}^{t, m}_\pi\), taken directly from \(H_t\). The top-right part represents the suitability of newly proposed candidate data as exemplars for previously selected data, estimated using Eq.~\ref{eq: estimate top-right}. Similarly, the bottom-left part represents the suitability of previously selected data as exemplars for newly proposed candidate data, estimated using Eq.~\ref{eq: estimate bottom-left}. 
The bottom-right section is filled with the median values of the other three sections.

We regard $M_t$ as a continuously decaying momentum term for historical information preserving.
Specifically, we first calculate $R_{t+1}^{i}$ by Eq.~\ref{eq: ap}. Then, we apply a weighted sum of $M_t$ and $R_{t+1}^{i}$ to recall the historical information as shown in Eq.~\ref{eq: history-aware iter},
\begin{equation}
\small
\label{eq: history-aware iter}
        R_{t+1}^{i} = \alpha_i \cdot M_t + (1-\alpha_i) \cdot (\beta \cdot R_{t+1}^{i} + (1-\beta) \cdot R_{t+1}^{i-1}) 
\end{equation}
where $\alpha_i = \lambda \cdot \alpha_{i-1}$ is the momentum coefficient with a decay rate of $\lambda$, and $\beta$ is the official AP damping rate \citep{cluster-ap}.
Finally, $A_{t+1}^{i}$ is calculated by Eq.~\ref{eq: ap}. All $\alpha$, $\lambda$ and $\beta$ are predefined hyperparameters. 

\subsection{Representativeness Scoring}
The individual representativeness score encapsulates the results of the exemplar election, reflecting both how willing other samples are to be represented by a specific sample and how unwilling the specific sample is to be represented by others. 
As explained earlier, the responsibility value \(R[i,k]\) indicates the suitability of \(x_k\) to serve as the exemplar for \(x_i\), while the availability value \(A[i,k]\) reflects the appropriateness of \(x_i\) selecting \(x_k\) as its exemplar. The combined value \((A+R)[i,k]\) represents the total evidence supporting \(x_i\)'s selection of \(x_k\) as its exemplar \citep{cluster-ap}. Thus, the sum of the \(k\)-th column of \(A+R\) can be interpreted as the total votes received by \(x_k\), and the sum of the \(i\)-th row of \(A+R\) represents the total votes cast by \(x_i\) for different samples. Defining \(Z = A + R\), the representativeness score of \(x_k\) is then computed using Eq.~\ref{eq: representation score}.

\begin{equation}
\small
\label{eq: representation score}
        s^k_{rep} = \sum_{i=1}^{|X'_{t+1}|}Z[i, k] - \sum_{i=1}^{|X'_{t+1}|}Z[k, i] + Z[k,k]
\end{equation}

\subsection{Integration of Diversity and Quality}
\label{method: combination}

Both data quality and data diversity are crucial for instruction tuning, yet existing methods often focus on one or address them sequentially. We combine quality and diversity scores in three ways, both preceded by min-max normalization (Eq.~\ref{eq: min-max}) to ensure scale consistency, where $s_q^k$ refers to the quality score of $x_k$, and $s^k_{rep}$ refers to the corresponding diversity score. 

\begin{equation}
\small
\label{eq: min-max}
\begin{aligned}
     {s'}^{k}_{rep} &= \frac{s^{k}_{rep} - \min\limits_{x_{i} \in {B}^m_{t}} s^{i}_{rep}}
        {\max\limits_{x_{i} \in {X'}_{t+1}} s^{i}_{rep} - \min\limits_{x_{i} \in {B}^m_{t}} s^{i}_{rep}},\quad \\
    {s'}_{q}^{k} &= \frac{s_{q}^{k} - \min\limits_{x_{i} \in {X'}_{t+1}} s_{q}^{i}}
        {\max\limits_{x_{i} \in {X'}_{t+1}} s_{q}^{i} - \min\limits_{x_{i} \in {X'}_{t+1}} s_{q}^{i}}
\end{aligned}  
\end{equation}

\begin{equation}
\small
\label{eq: addition-combine}
    s^{k} = {s'}^{k}_{rep} + \gamma \cdot {s'}_{q}^{k}.
\end{equation}

\begin{equation}
\small
\label{eq: multi-combine}
s^{k} =  (1+{s'}_{rep}^{k}) * (1+{s'}_{q}^{k})^{\gamma}
\end{equation}

Eq.~\ref{eq: addition-combine} and Eq.~\ref{eq: multi-combine} illustrate the calculation of the individual overall score using the additive and multiplicative approaches, respectively, where $\gamma$ is the weighting coefficient that controls the focus between diversity and quality.

In practice, we observe that further improving quality beyond a certain level can reduce the fine-tuned model's performance. Additionally, when combining quality and diversity using linear methods, diversity scores often dominate the selection process. This occurs because quality, as a linear score, increases at a constant rate, even when excessively large values provide diminishing benefits. More details can be found in our experimental analysis of score combination (Section~\ref{exp: combination}). 

\begin{table*}[htbp]
    \centering
    \small 
    \setlength{\tabcolsep}{5pt} 
    \begin{tabular}{l|ccc|ccc|ccc}
    \toprule
    \multirow{2}{*}{Method} & \multicolumn{3}{c|}{Llama3-8B} & \multicolumn{3}{c|}{Qwen2.5-7B} & \multicolumn{3}{c}{Mistral-7B} \\
    \cmidrule(lr){2-4} \cmidrule(lr){5-7} \cmidrule(lr){8-10}
           & AlpacaEval & MT-Bench & IFEval & AlpacaEval & MT-Bench & IFEval & AlpacaEval & MT-Bench & IFEval \\
    \midrule
    Full   & 19.07      & 5.88     & \underline{40.29} & 20.37   & 6.11    & 41.37    & 13.12   & 4.98   &   \textbf{35.25}  \\
    Random & 17.93      & 5.13     & 38.13             & 22.80   & 6.00    & 43.53    & 11.93    & 4.39   & 9.95     \\
    kCenter & 15.28     & 4.99     & 37.29             & 27.39   & 6.12    & \underline{46.40}    & 9.20  & 3.97    & 1.92     \\
    DEITA  & \underline{43.60} & 6.03 & 38.25         & \underline{50.43} & \underline{6.86} & 45.44 & \underline{28.82} & \underline{4.93} & 33.57 \\
    kNN$_1$ & 40.62     & \underline{6.04} & 38.49     & 46.96   & 6.62    & 45.56    & 26.62    & 4.91    & \underline{33.81} \\
    PIBE (ours)   & \textbf{44.84} & \textbf{6.23} & \textbf{40.89} & \textbf{51.55} & \textbf{6.88} & \textbf{46.76} & \textbf{29.48} & \textbf{5.03} & 29.38 \\
    \bottomrule
    \end{tabular}
    \caption{Comparison between different methods. For AlpacaEval and MT-Bench, we employ gpt-4o as annotator. The \textbf{bold} text indicates the best results, and the \underline{underlined} text represents the second-best results. The results of more base models can be found in Appendix~\ref{appendix: More Base Models}.}
    \label{tab: result-main}
\end{table*}

To address this, we design a nonlinear mapping function for quality scores, shown in Eq.~\ref{eq: nonlinear-combine}. Here, \(Q_p\) denotes the \(p\)-th percentile, \(r_l\) and \(r_h\) represent the lower and upper percentiles, \(S'_q\) refers to the scaled quality scores, and \(\sigma(\cdot)\) is the sigmoid function. The function, illustrated in Figure~\ref{fig: nonlinear_fn}, leverages the sigmoid's steepness in \((-2, 2)\) to enhance the distinguishability of scores within \([\tau_l, \tau_h]\), while flattening growth for scores above \(\tau_h\). Data below \(\tau_l\) are less considered, as such low-quality data are rarely selected into InsBank. Finally, we combine diversity with the nonlinear-mapped quality scores.

\begin{equation}
\small
\begin{aligned}
    \tau_l &= Q_{r_l}(S'_q) \\
    \tau_h &= Q_{r_h}(S'_q) \\
    c_{mul} &= 4 / (\tau_h - \tau_l) \\
    c_{sub} &= \tau_l + 2 / c_{mul} \\
    {s''}^k_q &= \sigma(({s'}^k_q - c_{sub}) * c_{mul})
\end{aligned}
\label{eq: nonlinear-combine}    
\end{equation}

After getting the overall scores, in addition to serving as the criterion for InsBank data selection, users can quickly select a smaller subset according to the data ranking to suit their own training budget.

\section{Experiment}
\subsection{Experimental Setup}

\textbf{Candidate Instruction Data}  We aggregate five instruction datasets for general instruction following capability: Self-Instruct \citep{collection-self-instruct}, Alpaca (GPT-4) \citep{collection-alpaca-gpt4}, Dolly \citep{collection-dolly}, ShareGPT\footnote{We filter out incomplete conversations.} \citep{collection-sharegpt} and WilzardLM (alpaca) \citep{collection-wizardlm}, resulting in a mixed dataset of 278k samples. The statistics of each dataset is presented in Table~\ref{tab: dataset-statistics}.

\textbf{Training and Evaluation} In this work, we fine-tune Llama3.2-1B, Llama3.2-3B, Llama3-8B \citep{model-llama3}, Qwen2.5-7B, Qwen2.5-14B \citep{llm-qwen2.5} and Mistral-7B \citep{llm-mistral} on the selected InsBank. Following DEITA \citep{ds-deita}, we set the size of InsBank to 6k for the convenience of subset evolution. We also experiment with InsBank size of 1k and 3k, and the results can be found in Appendix~\ref{appendix: More InsBank Budgets}. During training, we further restrict the trainable tokens and the number of conversation turns. We adopt AlpacaEval \citep{benchmark-alpaca-eval}, MT-Bench \citep{benchmark-mtbench} and IFEval \citep{benchmark-ifeval} for automatic model alignment performance evaluation. More details about training and evaluation can be found in Appendix~\ref{appendixs: hyperparameters}.

\textbf{Baselines} We compare proposed PIBE with the following baselines:
\begin{itemize}[nosep]
    \item \textbf{Full} Train model on all candidate data.
    \item \textbf{Random} Randomly select $m$ samples from all candidate data. 
    \item \textbf{kNN$_1$} Measure the diversity of one sample by its euclidean distance to the nearest neighbor (Eq.~\ref{eq: knn}). The diversity score is first normalized and then combine with the normalized quality score by $s_i = (1+kNN_1^i) * (1+{s'}_q^i)^{\gamma}$ for data selection.
    \item \textbf{kCenter Greedy} \citep{indiv_eval_coreset2} The original kCenter Greedy algorithm is shown in Alg.~\ref{alg:kcentergreedy}. We take $\operatorname*{min}_{x_j\in S_b}d(e(x_i),e(x_j))$ as the individual diversity score and combine it with quality score in the same manner of kNN$_1$.
    \item \textbf{DEITA} Traverse the instruction pool in descending order of quality scores and add a sample to the selected subset if its maximum cosine similarity with existing selected samples is below a threshold \citep{ds-deita}.
\end{itemize}

\subsection{Performance of SFT with InsBank}

Table~\ref{tab: result-main} compares the performance of LLM trained on subsets selected by different approaches. PIBE consistently outperforms the baselines on such benchmarks, showing the superiority of our data selection method. We further fine-tune Qwen2.5 7B \citep{llm-qwen2.5} and Mistral 7B \citep{llm-mistral} for robustness analysis, and the results exhibit the same trends, demonstrating that our method is effective across different models. We also report the quality and diversity of subsets selected by different methods in Table~\ref{tab: main-subset-statistics}. 
From the results of data selection, PIBE and DEITA demonstrate higher quality and diversity compared to kCenter and kNN. DEITA produces subsets with the highest quality, primarily because it prioritizes quality during the data selection process by traversing candidates in descending order of quality. In contrast, PIBE treats quality and diversity equally, enabling the subset to achieve the highest diversity while maintaining decent quality. 
From the perspective of downstream task performance, models fine-tuned with high-quality data (DEITA, PIBE) generally outperform those fine-tuned on low-quality data (kCenter, kNN). However, despite achieving the highest quality, DEITA’s downstream performance falls short of the more diverse PIBE, validating the importance of data diversity when the quality level is acceptable.

\subsection{Orderliness of InsBank}
\begin{figure}[hbtp]
\begin{center}
\includegraphics[width=0.9\columnwidth]{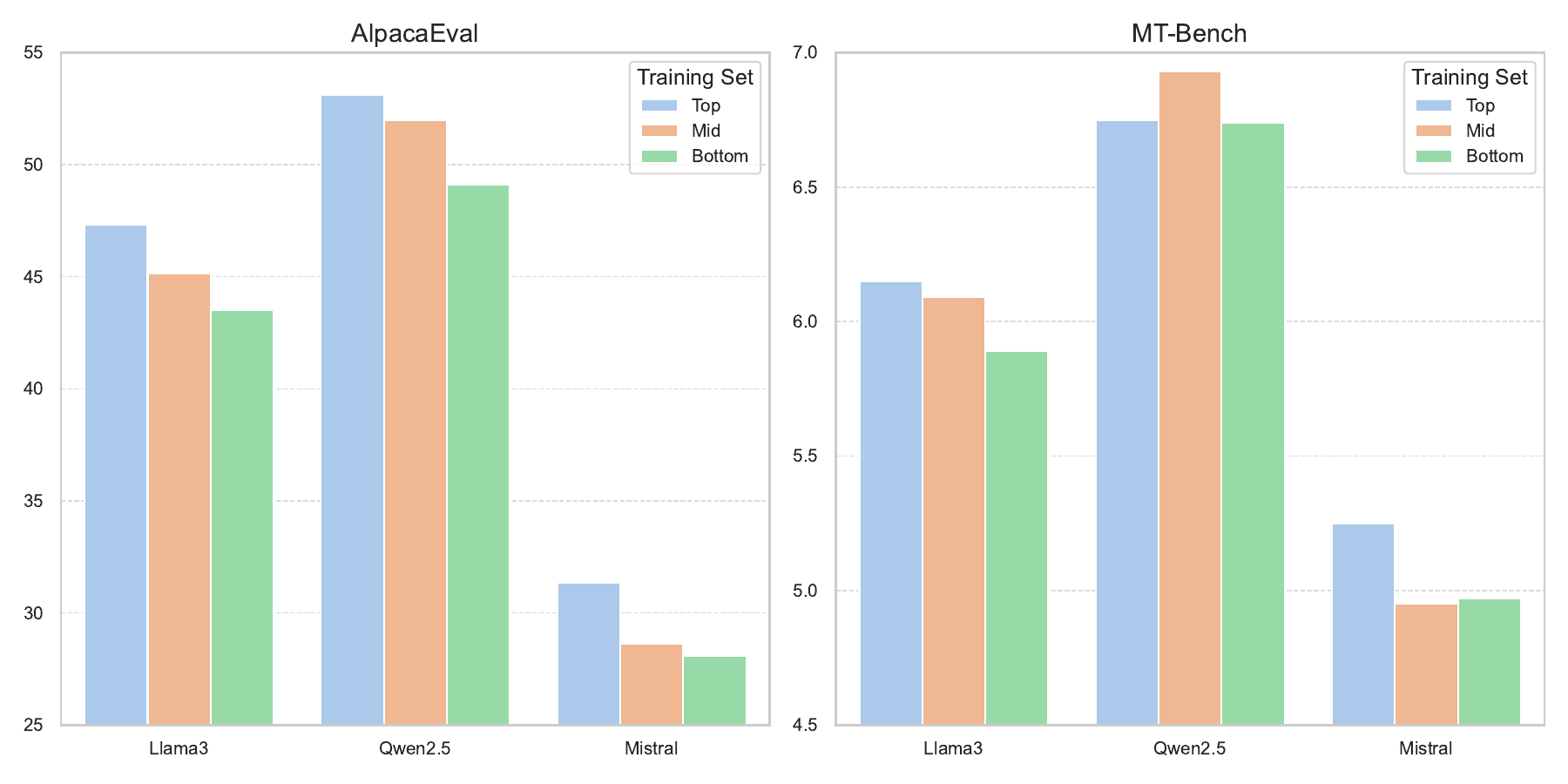}
\end{center}
\caption{Results of orderliness experiment.}
\label{fig: results-ordering}
\end{figure}

Each sample in the InsBank selected by PIBE is provided with an overall individual score reflects both the diversity and quality which shows the priority of each sample to be used to fine-tune models. We sort the InsBank in descending order based on the overall individual score, and compare the performance of models fine-tuned with the “top2k, mid2k, bottom2k” samples in InsBank. Here, we use the instruction subset obtained from the final evolution round, and restrict the trainable tokens to 0.9M and turns to 2.3k. The results are illustrated in Fig~\ref{fig: results-ordering}, showing that the top-ranked data generally achieved better performance, proving the orderliness of InsBank. 

\begin{table}[htbp]
    \centering
    \small
    \begin{tabular}{lcccc}
    \toprule
     Metric & kCenter & DEITA &  kNN$_1$ & PIBE \\
    \midrule
    Quality & 4.37 & \textbf{5.19} & 4.82 & 5.13 \\
    Diversity & 62.26 & 86.94 & 77.24 & \textbf{91.84} \\
    \bottomrule
\end{tabular}
\caption{The quality and diversity of subsets selected by different methods. The diversity here is measured by euclidean distance between data.}
\label{tab: main-subset-statistics}
\end{table}

\subsection{Analysis}

In this section, we analyze the effectiveness of diversity and quality. We also experiment PIBE with different score combination methods. More analysis about overlap between progressive evolving and full data selection, InsBank evolution, PIBE hyper-parameters, time costs and selected data quality distribution can be found in Appendix~\ref{appendix: addition-analysis}.

\textbf{Effectiveness of Diversity and Quality} To validate the role of diversity in instruction data selection, we first construct a quality-controlled subset where all data have quality scores within the range of 4.5 to 5.0 (details in Appendix~\ref{appendix: qc construction}). Using PIBE, we compute individual diversity scores for the subset, sort the data in descending order, and select the top 6k samples as the most diverse subset and the bottom 6k as the least diverse subset. The distributions of the two subsets are shown in Fig.~\ref{fig: qc_diversity}. Before fine-tuning, we restrict the total trainable tokens to 2M. Results in Table~\ref{tab: qc_diversity} indicate that, with comparable quality, models trained on more diverse data achieve better performance.

\begin{table}[htbp]
    \centering
    \small
    \begin{tabular}{l|cc|cc}
    \toprule
    Method  & Qua & Div & AlpacaEval & MT-Bench \\  
    \midrule
    Top & 4.84 & 81.14 & \textbf{27.70} & \textbf{5.52} \\
    Bottom & 4.86 & 68.55 & 27.33 & 5.43 \\
    \bottomrule
    \end{tabular}
    \caption{The results of quality-controlled diversity effectiveness experiment. Qua refers to the average quality score, and Div refers to the average diversity score.}
    \label{tab: qc_diversity}
\end{table}

When it comes to quality, the improvement from extremely low to high quality is clearly beneficial, as extremely low-quality subsets often contain noisy data, such as irrelevant or incomplete responses. However, \emph{is continuously improving quality always effective in the data selection process?} To address this, we compare model performance fine-tuned on data selected by the following strategies in the final evolution iteration: (1) \textbf{Diversity Greedy}: selecting data with the highest diversity scores; (2) \textbf{Quality Greedy}: selecting data with the highest quality scores; and (3) \textbf{PIBE}. The results shown in Table~\ref{tab: analysis_div_and_qua} reveal a clear trade-off between diversity and quality. A purely greedy approach focusing on either aspect leads to suboptimal outcomes, while a balanced consideration of both proves more effective. This finding aligns with the main experiment results and suggests the existence of a balance point between diversity and quality, which we further investigate through the analysis of score combination.

\begin{table}[htbp]
    \centering
    \small
    \begin{tabular}{l|cc|cc}
    \toprule
    Method  & Qua & Div & AlpacaEval & MT-Bench \\  
    \midrule
    DG & 5.02 & 93.06 & \textbf{41.93}  & \textbf{6.09} \\
    QG & 5.20 & 83.70 & 40.86  & 5.86 \\
    PIBE & 5.13 & 91.84 & 44.84 & 6.23 \\
    \bottomrule
    \end{tabular}
    \caption{Analysis of diversity and quality contribution. Here, DG refers to diversity greedy, and QG refers to quality greedy}
    \label{tab: analysis_div_and_qua}
\end{table}

\label{exp: combination}
\textbf{Analysis of Score Combination} We experiment with the different combination methods to explore the contribution of quality and diversity in PIBE. 

\begin{table}[htbp]
    \centering
    \small
    \setlength{\tabcolsep}{2pt} 
    \begin{tabular}{l|cc|ccc}
    \toprule
    Param & AlpacaEval & MT-Bench & SP-Qua & SP-Div & Diff \\
    \midrule
    \multicolumn{6}{c}{Multiplication} \\
    \midrule
    $\gamma=1$    & 44.84 & 6.23 & 0.36 & 0.74 & 0.38 \\
    $\gamma=2$    & \textbf{46.77} & \textbf{6.15} & 0.51 & 0.70 & 0.19 \\
    $\gamma=3$    & 42.98 & 6.17 & 0.54 & 0.67 & 0.13 \\
    \midrule
    \multicolumn{6}{c}{Addition} \\
    \midrule
    $\gamma=1$    & 44.84 & 6.13 & 0.44 & 0.72 & 0.28 \\
    $\gamma=2$    & \textbf{47.08} & \textbf{6.10} & 0.54 & 0.68 & 0.14 \\
    $\gamma=3$    & 44.53 & 6.09 & 0.56 & 0.64 & 0.08 \\
    \midrule
    \multicolumn{6}{c}{Nonlinear} \\
    \midrule
    $r_h=0.80$ & 44.41 & 5.98 & 0.58 & 0.72 & 0.14 \\
    $r_h=0.90$ & 44.84 & 6.19 & 0.62 & 0.70 & 0.08 \\
    $r_h=0.95$ & \textbf{47.58} & \textbf{6.36} & 0.63 & 0.69 & 0.06 \\
    \bottomrule
\end{tabular}
\caption{The results of different combination methods. SP- refers to Spearman value, Diff refers to the difference value between SP-Qua and SP-Div.}
\label{tab: result-combination}
\end{table}

We first explore the multiplication manner and the addition manner, and the results are reported in Table~\ref{tab: result-combination}. Overall, regardless of whether addition or multiplication is used as the combination method, the results exhibit a distinct trend of initially increasing and then decreasing as the influence of quality grows (i.e., with the increase of the $ \gamma $ value). This finding supports the hypothesis that a balance point exists between diversity and quality.

We analyze the correlation between quality and selection flags, as well as diversity and selection flags, for the top 12k data sorted by overall score (details in Appendix~\ref{appendix: correlation-analysis}). As shown in Table~\ref{tab: result-combination}, Spearman for diversity consistently surpass those for quality, indicating diversity's priority during selection. While increasing \(\gamma\) reduces the gap, this approach presents limitations: (1) Even at \(\gamma=3\), a notable gap remains between SP-Qua and SP-Div, particularly with the multiplication method; (2) Increasing \(\gamma\) further improves downstream performance initially but leads to declines afterward. 

Examining the quality distribution of selected data (Figure~\ref{fig: selected-data-quality-distribution}), we observe that \(\gamma=1\) includes some low-quality data, while \(\gamma=3\) selects excessive high-quality data. As discussed in Section~\ref{method: combination}, this stems from quality's linear nature. To address this, we use a nonlinear quality mapping function. Fixing \(r_l=0.3\), we compare different \(r_h\) values, with results shown in Table~\ref{tab: result-combination}. Nonlinear mapping significantly mitigates diversity's dominance and improves fine-tuned model performance, particularly at \(r_h=0.95\). Unlike linear methods, which improve subset quality by selecting extreme high-quality values, the nonlinear approach raises overall quality by incorporating more moderately high-quality data, aligning with its design goals.

\section{Related Work}
Instruction fine-tuning is widely used to refine LLMs. Early methods focused on fine-tuning with large-scale instruction datasets \citep{collection-flan-2021, collection-super-natural-inst} manually aggregated from extensive NLP task collections \citep{collection-flan-2022}. With advancements in generative models, \citet{collection-self-instruct} has led the trend of synthetic data generation \citep{collection-alpaca, collection-ultrachat, collection-wizardlm}. 
As \citet{ift-lima} found, quality and diversity are more important than quantity, driving recent efforts to cut training costs by removing low-quality and redundant data.
Existing selection methods can be broadly categorized into three types \citep{ds-survey}: quality-based, diversity-based, and model-specific importance-based selection.

\textbf{Quality-based Selection} 
Humpback \citep{ds-self-alignment} selects high-quality samples through an iterative self-curation process where quality predictions are produced by the fine-tuned model of each turn. 
Recent works typically employ a GPT-model to annotate the data quality.  For example, ALPAGASUS \citep{ds-alpagasus} employs ChatGPT to score the accuracy of instruction data and select data according to a threshold. 

\textbf{Diversity-based Selection} 
The diversity-based selection aims to deduplicate the instruction data and maximize the coverage of selected data. Recent methods typically achieve this purpose by control the nearest neighbor distance \citep{ds-deita} or maximize the average distance between the selected data through text embedding \citep{ds-self-evolve}. INSTAG \citep{ds-instag} identifies semantics and intentions of instructions by tags and it assumes that a dataset is considered more diverse if it covers more individual tags.

\textbf{Model-specific Importance-based Selection} Importance refers to the necessity of adding one sample into training set \citep{ds-survey} whose indicator are typically model-specific \citep{ds-less, ds-ifd, ds-ic-ifd, ds-mods}. However, this work focuses on the general data selection and emphasizes the quality and diversity of selected data.

InfoGrowth \citep{data-growth} also aims to address the continuous expansion of datasets, but it primarily focuses on image data and relabeling noisy samples, making it less relevant to this paper. While InfoGrowth and DEITA consider both quality and diversity, they handle them sequentially, without combining them into a unified score. Besides, previous efforts primarily aggregate all candidate data before data selection and are not experimented under the progressive instruction bank evolution task. In this paper, we propose PIBE to efficiently obtain the optimal current instruction subset with comprehensive characterization and integration of diversity and quality scores.

\section{Conclusion}

In this paper, we propose InsBank to address the challenge of evolving instruction subset. PIBE integrates high-quality and representative data into InsBank, striking a balance between data diversity and quality, while maintaining long-term scalability and efficiency. By leveraging a representation-based diversity score with historical information, PIBE flexibly combines diversity and quality for data selection and ranking. Experimental results show PIBE outperforms baselines, providing more optimal and adaptable instruction subsets. The orderliness of InsBank also allows users to extract tailored subsets within budget constraints, supporting cost-effective training and the ongoing refinement of LLMs. This work paves the way for more dynamic and adaptable instruction tuning strategies, enhancing both the efficiency and effectiveness of LLM development over time.
\section*{Limitations}

In this work, we focus on evaluating the diversity of individual instruction data and exploring the combination of diversity and quality scores. However, achieving a more precise assessment of data quality remains a valuable direction for future research.

\section*{Ethics Statement}
All of the datasets used in this study were publicly available, and no annotators were employed for our data collection. We confirm that the datasets we used did not contain any harmful content and was consistent with their intended use (research). We have cited the datasets and relevant works used in this study.

\bibliography{acl_latex}

\newpage

\appendix

\section{Timeline of Instruction Datasets}
\label{appendix: timeline}
\begin{figure}[htbp]
\begin{center}
\includegraphics[width=0.8\columnwidth]{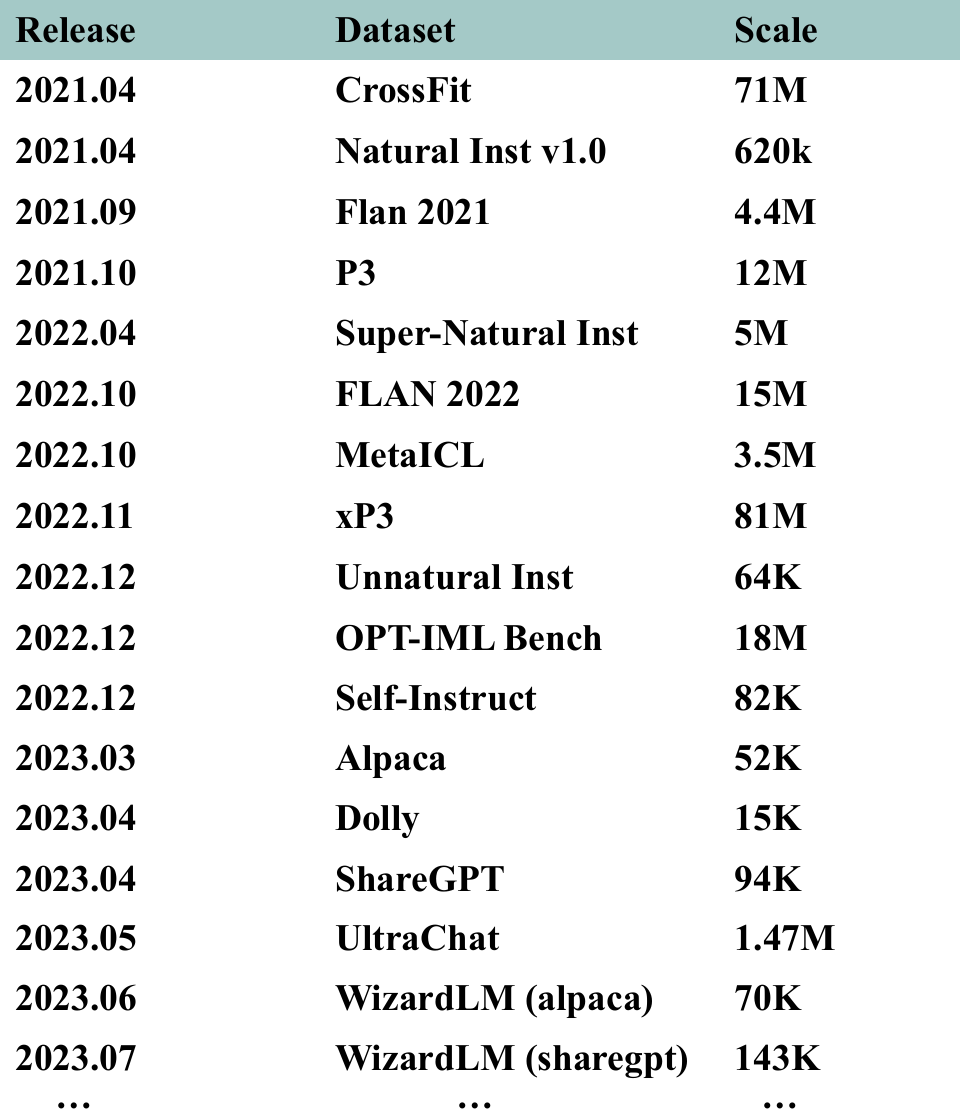}
\end{center}
\caption{Timeline of instruction datasets (part) since 2021.04 to 2023.07.} 
\label{fig: timeline}
\end{figure}

\section{Momentum Responsibility Matrix}
\label{appendix: momentum-responsibility-matrix}

\begin{figure}[hbtp]
\includegraphics[width=\columnwidth]{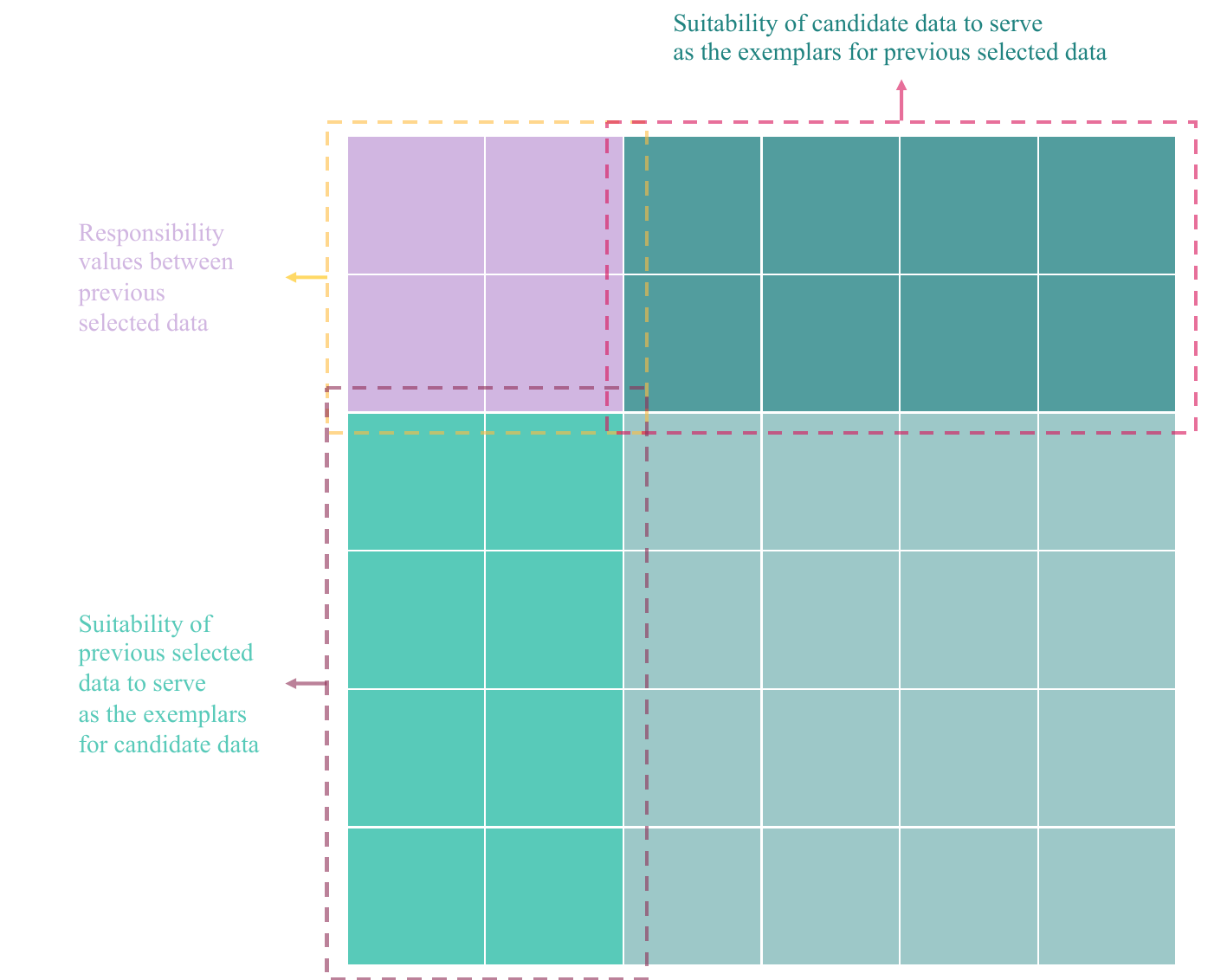}
\caption{The structure of momentum responsibility matrix.}
\label{fig: historical information}
\end{figure}

\section{Details of Implementation}
\label{appendixs: hyperparameters}
\textbf{Fine-grained Quality Scoring} We adopt the quality annotator \footnote{\url{https://huggingface.co/hkust-nlp/deita-quality-scorer}} provided by \citet{ds-deita} to score the instructions.

\textbf{Representation-based Progressive Data Selection:} During the PIBE data selection process, we set the momentum coefficient $\alpha=0.3$, the momentum decaying rate $\lambda=0.9$, the damping rate $\beta=0.5$ and the weighting coefficient $\gamma=1$. Besides, we adopt instruction embedding \citep{instruction_embedding} to encode the instructions. As for affinity propagation, we use negative euclidean distance to initialize the similarity matrix and fill the diagonal of similarity matrix with 0. Moreover, due to the high memory overhead of Affinity Propagation (\(O(n^3)\)), we further divided the complete set of candidates in each evolution iteration into smaller evolution batches with a batch size of 27,000 to perform PIBE. For data selection, all baselines employ the full-scale selection manner rather than the gradual selection manner to get their global optimal performance. For PIBE, we perform progressive InsBank evolution following the temporal order of dataset appearance (i.e. Self-Instruct $\rightarrow$ Alpaca $\rightarrow$ Dolly $\rightarrow$ ShareGPT $\rightarrow$ WizardLM), and take the final selected subset for model fine-tuning.

\textbf{Instruction Fine-Tuning:} We utilize 8 NVIDIA A100 SXM4 40GB GPUs to fine-tune LLMs. We employ LlamaFactory \citep{llamafactory}, DeepSpeed Zero-Stage 3 \citep{deepspeed} and fp16 precision to facilitate the training process. We adopt the Llama3-style template for Llama3-8B, Qwen-style template for Qwen2.5-7B and Mistral-style template for Mistral-7B, corresponding to "llama3" "qwen," and "mistral" template in LlamaFactory respectively. We set the effective batch size to 128 (per device train batch size=1 and gradient accumulation steps=16), training epochs to 6, learning rate to 1e-5, warmup ratio to 0.1 and maximum input length to 2048.

For trainable tokens and turns restriction, we set max tokens to 3M and max turns to 7k unless otherwise specified. For quality-controlled experiments, since all data are single-turn conversations, we set max tokens to 2M and max turns to 6k. For orderliness analysis, we set max tokens to 0.9M and max turns to 2.3k.

For AlpacaEval inference, we set temperature=0.7, top\_p=0.9, top\_k=40, num beams=1 and max length=512. For MT-Bench inference, we follow the default setting of FastChat\footnote{\url{https://github.com/lm-sys/FastChat/tree/main}} except for that max length is set to 512. All models adopt templates consistent with those in the training process during evaluation.

For AlpacaEval evaluation, we compare each model output with GPT-3.5 Turbo (gpt-3.5-turbo-1106) \citep{openai_chatgpt}, because we find that when compared to text-davinci-003 \citep{gpt3} or GPT-4 Turbo \citep{gpt4}, the benchmark was either too simple or too challenging, making it difficult to differentiate between models. For both AlpacaEval and MT-Bench, we employ GPT-4o \citep{openai_gpt4o} as annotator.

\section{Correlation Analysis}
\label{appendix: correlation-analysis}

We first sort the data in descending order based on the overall score and select the top 12k samples. For each sample, we assign a flag: if the sample is selected into InsBank, the flag is set to 1; otherwise, it is set to 0. We then calculate the Spearman correlation coefficients between diversity and flags, as well as between quality and flags, to investigate the contributions of diversity and quality to data selection. We restrict our analysis to the top 12k data sorted in descending order by the overall score, as we aim to focus on high-quality candidates with relatively high quality and diversity. Lower-quality candidates are excluded from the analysis since their likelihood of being selected into InsBank is inherently low.

\section{Quality-Controlled Subset Construction}
\label{appendix: qc construction}

To avoid mixing single-turn and multi-turn conversations data, as well as biases introduced by different data distributions across dataset, we sample data with quality ranging from 4.5 to 5.0 from WizardLM (alpaca), resulting in a quality-controlled subset with 19805 samples. 

\section{Selected Data Visualization from QC-Subset}

\begin{figure}[htbp]
\begin{center}
\includegraphics[width=\columnwidth]{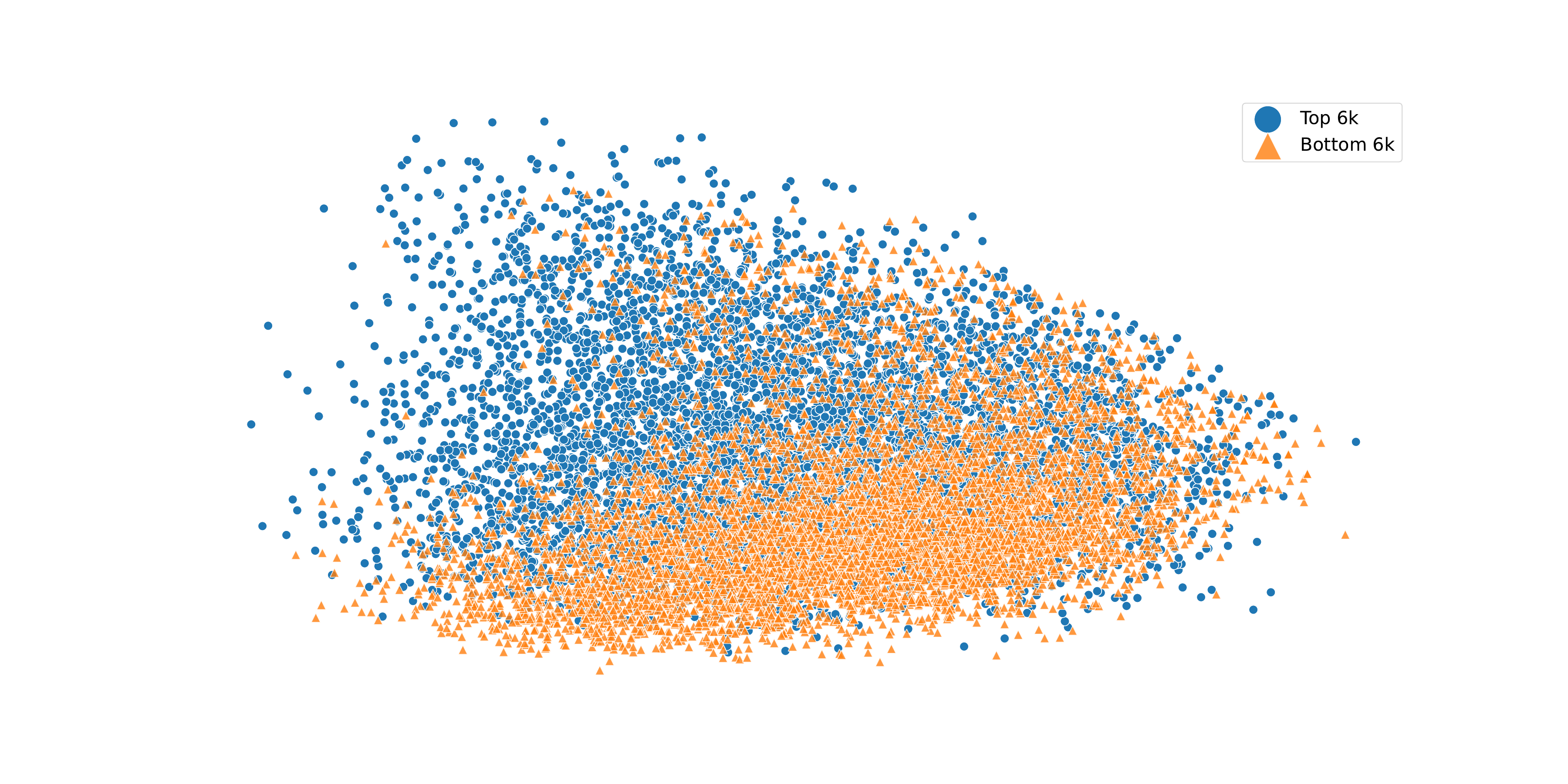}
\end{center}
\caption{Selected data visualization based on quality controlled subset. The blue stars represent the most diverse data, while the orange triangles represent the least diverse data.}
\label{fig: qc_diversity}
\end{figure}

\section{Statics of Candidate Instruction Datasets}

\begin{table}[H]
    \centering
    \small
    \begin{tabular}{lcc}
    \toprule
    Dataset & Scale & Quality \\
    \midrule
    Self-Instruct & 82k & 2.29 \\
    Alpaca & 52k & 3.59 \\
    Dolly & 15k & 2.76 \\
    ShareGPT (cleaned) & 58k & 4.03 \\
    WizardLM & 70k & 4.16 \\
    \bottomrule
    \end{tabular}
    \caption{Statistics of instruction datasets.}
    \label{tab: dataset-statistics}
\end{table}

\section{Nonlinear Quality Mapping Function}

\begin{figure}[hbtp]
\includegraphics[width=\columnwidth]{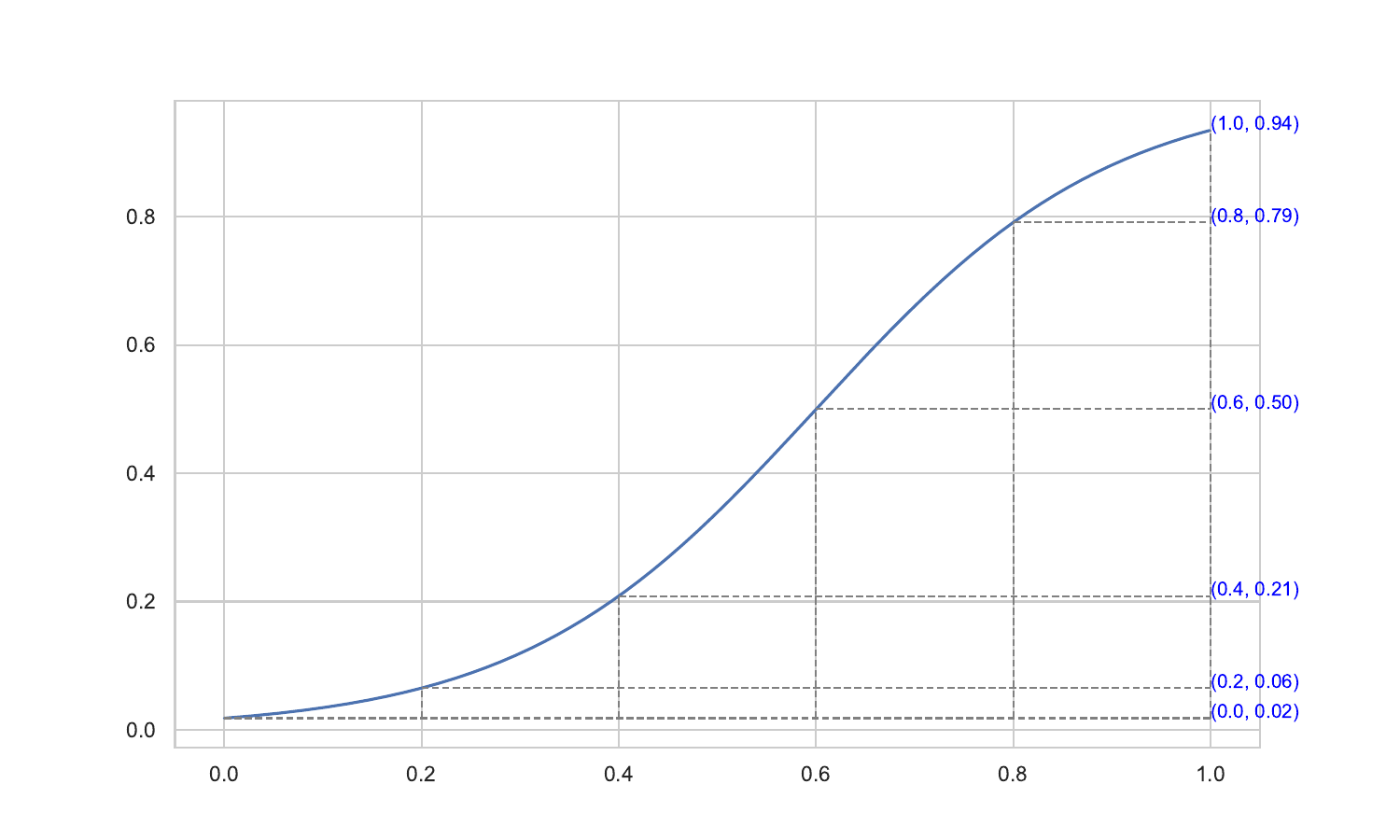}
\caption{Visualization of nonlinear quality mapping function.}
\label{fig: nonlinear_fn}
\end{figure}

\section{K-Center Greedy Algorithm}

\begin{algorithm}
\caption{K-Center Greedy}\label{alg:kcentergreedy}
\begin{algorithmic}[1]
\Require data $x_i\in S$ and a budget $m$
\State Initialize $S_m=x_0$
\Repeat
\State $u=\arg\operatorname*{max}_{x_i\in S\backslash S_m}$
\Statex
\quad\quad\quad\quad\quad\quad\quad\quad$\operatorname*{min}_{x_j\in S_m}d(g(x_i),g(x_j))$
\State $S_m = S_m\cup \{u\}$
\Until{$|S_{m}|=m$}
\State \Return {$S_{m}$}
\end{algorithmic}
\end{algorithm}

\section{Additional Analysis}
\label{appendix: addition-analysis}


\subsection{Justification of Data Composition}
\label{appendix: Justification of Data Composition}

The data composition in this work simulates the development of instruction sets. Although Self-Instruct, Alpaca, and WizardLM are related to each other, their instruction data are actually different from each other. In addition, the balance between quality and diversity during data selection is also one of the key focus of this work. By utilizing candidate instruction sets with varying quality distributions, we demonstrate that PIBE is capable of jointly considering both quality and diversity. 

In this work, we focus on the efficient instruction subset evolution during the development of instruction data, thus we select Self-Instruct, Alpaca (GPT-4), Dolly, ShareGPT, and WizardLM as candidate instruction sets based on their chronological release order. These datasets collectively exhibit a trend of increasing data quality which aligns well with our scenario of data evolution. 

Additionally, both quality and diversity are essential to data selection, and we have demonstrated in this paper that solely focus on one underperforms comprehensively consider both (Table \ref{tab: analysis_div_and_qua}). Therefore, high quality data alone are far from enough and including data of moderate quality to enhance data diversity is of great value. We report the quality and diversity of subsets selected by different methods in Table~\ref{tab: main-subset-statistics} and Table~\ref{tab: more_baselines-2}, showing that PIBE is able to maintain decent data quality while achieve the highest level of diversity against MoDS and DEITA. Moreover, the InsBank data distribution of each evolution step is also shown in Table~\ref{tab: data_source}. The final InsBank mainly consists of data from high quality datasets (ShareGPT, WizardLM), while some data from medium quality dataset (Alpaca) are also included to further enhance the diversity of InsBank. Only a limited number of samples from low-quality datasets (Self-Instruct, Dolly) are present in InsBank, showing that PIBE is able to effectively ignore low-quality samples during evolution.

\begin{table}[htbp]
    \centering
    \small
    \begin{tabular}{ccccc}
    \toprule
     Self-Instruct & Alpaca & Dolly & ShareGPT & Wizard \\
    \midrule
    6000 & - & - & - & - \\
    144 & 5856 & - & - & - \\
    114 & 5695 & 192 & - & - \\
    9 & 1832 & 17 & 4142 & - \\
    3 & 632 & 17 & 2177 & 3181 \\
    \bottomrule
\end{tabular}
\caption{InsBank composition in different stage of InsBank evolution.}
\label{tab: data_source}
\end{table}

\subsection{Effectiveness of Data Selection}
\label{appendix: Effectiveness of Data Selection}

To better demonstrate the effectiveness of data selection with high quality data, we first randomly sampled 50k data from the high quality dataset - UltraChat\citep{collection-ultrachat}. Then, we perform DEITA and PIBE to select a 6k subset from it separately. We compare the performance of model fine-tuned with Full, DEITA and PIBE, and the results are shown in Table~\ref{tab: ultrachat}. Both DEITA and PIBE outperform the full-data baseline, further confirming the benefits of appropriate data selection for model instruction fine-tuning. Notably, PIBE achieves the best performance, which further demonstrates its superiority.

\begin{table}[htbp]
    \centering
    \small
    \begin{tabular}{lccc}
    \toprule
    Method  & AlpacaEval & MT-Bench & IFEval \\  
    \midrule
    Full & 20.27 & 5.12 & 32.01\\
    DEITA & 24.75 & 5.64 & 30.82 \\
    PIBE & 27.86 & 5.73 & 29.26 \\
    \bottomrule
    \end{tabular}
    \caption{Data selection performance with UltraChat.}
    \label{tab: ultrachat}
\end{table}

\subsection{More Baselines for Comparison}
\label{appendix: More Baselines for Comparison}

In this section, we further compare PIBE with three model-specific baselines: IFD\citep{ds-ifd}, IC-IFD\citep{ds-ic-ifd} and MoDS\citep{ds-mods}. As shown in Table~\ref{tab: more_baselines-1} and Table~\ref{tab: more_baselines-2}, PIBE consistently outperforms these baselines attributing to its better balance between data quality and data diversity. For the underperformance of IFD and IC-IFD, it may due to the fact that the IFD-style metric does not guarantee the high quality of the selected data. We further check the average quality scores of subsets selected by IFD and IC-IFD, and they are significantly lower than DEITA and PIBE. For MoDS, it greatly outperforms IFD, IC-IFD due to its quality filtering strategy. However, it also handles data quality and data diversity separately, making them less balanced during the data selection process. Moreover, the augmented data selection process of MoDS is also time-consuming, making it less efficient than other data selection methods. 

\begin{table}[htbp]
    \centering
    \small
    \begin{tabular}{lccc}
    \toprule
    Method  & AlpacaEval & MT-Bench & IFEval \\  
    \midrule
    IFD & 24.50 & 5.01 & 36.57 \\
    IC-IFD & 30.04 & 5.57 & 36.45 \\
    MoDS & 42.83 & 5.83 & 38.01 \\
    PIBE & \textbf{44.84} & \textbf{6.23} & \textbf{40.89} \\
    \bottomrule
    \end{tabular}
    \caption{Results of comparison between PIBE and model-specific baselines.}
    \label{tab: more_baselines-1}
\end{table}

\begin{table}[htbp]
    \centering
    \small
    \begin{tabular}{lcccc}
    \toprule
     Metric & IFD & IC-IFD & MoDS & PIBE \\
    \midrule
    Quality & 3.44 & 3.54 & 5.20 & 5.13 \\
    Diversity & 111.82 & 117.38 & 82.51 & 91.84 \\
    \bottomrule
\end{tabular}
\caption{The quality and diversity of subsets selected by different methods. }
\label{tab: more_baselines-2}
\end{table}

\subsection{More InsBank Budgets}
\label{appendix: More InsBank Budgets}

\begin{table}[htbp]
    \centering
    \small
    \begin{tabular}{lccc}
    \toprule
    Method  & AlpacaEval & MT-Bench & IFEval \\  
    \midrule
    \multicolumn{4}{c}{\textbf{Budget=1k}} \\
    \midrule
    DEITA & 13.06 & 4.53 & \textbf{37.17} \\
    PIBE  & \textbf{20.77} & \textbf{4.69} & 34.05 \\
    \midrule
    \multicolumn{4}{c}{\textbf{Budget=3k}} \\
    \midrule
    DEITA & \textbf{43.15} & 5.71 & 38.97 \\
    PIBE  & 42.79 & \textbf{5.90} & \textbf{39.33} \\
    \midrule
    \multicolumn{4}{c}{\textbf{Budget=6k}} \\
    \midrule
    DEITA & 40.62 & 6.23 & 38.49 \\
    PIBE  & \textbf{44.84} & \textbf{6.23} & \textbf{40.89} \\
    \bottomrule
    \end{tabular}
    \caption{Results of InsBank budget of 1k and 3k.}
    \label{tab: more-budget}
\end{table}

We further conduct experiments with InsBank budget of 1k and 3k. The results in Table~\ref{tab: more-budget} show that increasing the data size from 1k to 3k leads to a significant improvement in model performance. However, when the data size is further increased from 3k to 6k, the performance gain becomes relatively marginal. This reflects a trend in which the model’s general instruction-following ability improves rapidly with more training data but also converges quickly, which is consistent with the observations reported in \citep{how-abilities}.

\subsection{More Base Models}
\label{appendix: More Base Models}

We further conduct experiments on Llama3.2-1B, Llama3.2-3B \citep{model-llama3}, Qwen2.5-14B \citep{llm-qwen2.5}, the results shown in Table~\ref{tab: more-base-models} indicate that models of sizes 1B, 3B, and 14B all greatly benefit from the instruction data and our experimental findings can further generalize to models of sizes 1B, 3B, and 14B.

\begin{table}[htbp]
    \centering
    \small
    \begin{tabular}{lccc}
    \toprule
    Method  & AlpacaEval & MT-Bench & IFEval \\   
    \midrule
    \multicolumn{4}{c}{\textbf{Qwen2.5-14B}} \\
    \midrule
    base & 12.19 & 6.89 & 41.01 \\
    DEITA & 58.40 & 7.34 & 45.32 \\
    PIBE & \textbf{58.58} & \textbf{7.46} & \textbf{46.52} \\
    \midrule
    \multicolumn{4}{c}{\textbf{Qwen2.5-7B}} \\
    \midrule
    base & 14.68 & 6.61 & 40.05 \\
    DEITA & 50.43 & 6.86 & 45.44 \\
    PIBE & \textbf{51.55} & \textbf{6.88} & \textbf{46.76} \\
    \midrule
    \multicolumn{4}{c}{\textbf{Llama3-8B}} \\
    \midrule
    base & 0.75 & 2.03 & 20.14 \\
    DEITA & 43.60 & 6.03 & 38.25 \\
    PIBE & \textbf{44.84} & \textbf{6.23} & \textbf{40.89} \\
    \midrule
    \multicolumn{4}{c}{\textbf{Llama3.2-3B}} \\
    \midrule
    base & 0.49 & 1.56 & 17.99 \\
    DEITA & 29.73 & 4.71 & 36.33 \\
    PIBE & \textbf{29.98} & \textbf{4.96} & \textbf{38.85} \\
    \midrule
    \multicolumn{4}{c}{\textbf{Llama3.2-1B}} \\
    \midrule
    base & 0.00 & 1.10 & 17.99 \\
    DEITA & \textbf{8.96} & 3.28 & 31.65 \\
    PIBE & 8.21 & \textbf{3.39} & \textbf{31.77} \\
    \bottomrule
    \end{tabular}
    \caption{Results of further experiment with different base models.}
    \label{tab: more-base-models}
\end{table}

\begin{table*}[htbp]
    \centering
    \small
    \begin{tabular}{lcccccc}
    \toprule
    Method  & MMLU & HellaSwag & ARC & TruthfulQA & Winogrande & Avg \\  
    \midrule
    Full    & 58.47 & 79.20 & 55.03 & 50.06 & 73.32 & 63.21 \\
    Random  & 60.34 & \textbf{83.39} & 57.88 & 44.69 & 71.88 & 63.63 \\
    kCenter & 62.00 & 80.97 & 58.79 & 44.97 & 72.77 & 63.89 \\
    kNN     & \textbf{64.29} & 82.41 & 59.04 & 52.74 & 74.03 & 66.50 \\
    DEITA   & 64.15 & 82.95 & 59.90 & 51.81 & 74.43 & 66.64 \\
    PIBE    & 63.76 & 82.38 & \textbf{61.18} & \textbf{53.55} & \textbf{75.37} & \textbf{67.24} \\
    \bottomrule
    \end{tabular}
    \caption{Results of OpenLLM evaluation}
    \label{tab: more_benchmarks-1}
\end{table*}

\begin{table*}[htbp]
    \centering
    \small
    \begin{tabular}{l|ccccc|ccccc|c}
    \toprule
    \multirow{2}{*}{Method} & \multicolumn{5}{c|}{HSR} & \multicolumn{5}{c|}{SSR} & CSL \\
    \cmidrule(lr){2-6} \cmidrule(lr){7-11} \cmidrule(lr){12-12}
    \noalign{\vskip 3pt}
    & L1 & L2 & L3 & L4 & L5 & L1 & L2 & L3 & L4 & L5 & CSL \\  
    \midrule
    DEITA & 43.26 & 47.18 & 36.97 & 22.41 & 26.46 & 43.26 & 59.96 & 51.37 & 46.03 & 48.99 & 1.12 \\
    PIBE  & \textbf{59.00} & \textbf{53.86} & \textbf{42.79} & \textbf{30.36} & \textbf{31.93} & \textbf{59.00} & \textbf{63.12} & \textbf{57.26} & \textbf{48.98} & \textbf{56.73} & \textbf{1.62} \\
    \bottomrule
    \end{tabular}
    \caption{Results of FollowBench evaluation}
    \label{tab: more_benchmarks-2}
\end{table*}

\subsection{Further Evaluation with More Benchmarks}
\label{appendix: urther Evaluation with More Benchmarks}

We further extend the main experiments with Llama3-8B to more benchmarks (MMLU \citep{benchmark-mmlu}, HellaSwag \citep{benchmark-hellaswag}, ARC \citep{benchmark-arc}, TruthfulQA \citep{benchmark-truthfulqa}, Winogrande \citep{benchmark-winogrande} and FollowBench \citep{benchmark-followbench}), and the results are shown in Table~\ref{tab: more_benchmarks-1} and Table~\ref{tab: more_benchmarks-2}. PIBE consistently outperforms all baselines in the further evaluations, demonstrating its overall superiority.

\subsection{Overlap Between Progressive Evolving and Full Data Selection}
\label{appendix: overlap-between-progressive-and-full}

In this section, we aim to compare the overlap rates between the subsets selected by different methods from the gradual manner and those from the full-scale selection manner \footnote{Aggregate all available candidates first and perform data selection on the full data directly.}. 

We randomly select 40k data from the full data to obtain a subset that closely resembles the distribution of real data. We set the InsBank size here to 1k, and divided the data into four candidate subsets of 10k each to simulate the gradual manner. We compared PIBE with kNN$_1$ and k-Center Greedy, and perform an ablation analysis on the historical information used in PIBE. We set $\gamma=1$, and for PIBE, we set $\alpha=0.3$ and $\lambda=0.9$ which aligns with the main experiment. The results are reported in Table~\ref{tab: result-overlap}. It shows that the overlap rate of PIBE exceeds that of the kNN$_1$ and kCenter Greedy, and the historical information also helps improve the overlap rate. 

\begin{table}[htbp]
    \centering
    \small
    \begin{tabular}{lcccc}
    \toprule
    Method  & k-NN & kCenter & PIBE w/o hst & PIBE \\  
    \midrule
    Num & 131 & 747 & 390 & 864 \\
    \bottomrule
    \end{tabular}
    \caption{The overlap sample number between subset selected in full-scale manner and in gradual manner. Here, PIBE w/o hst is the ablation on history information of PIBE.}
    \label{tab: result-overlap}
\end{table}

\subsection{Instruction Bank Evolution}


\begin{figure}[htbp]
  \includegraphics[width=\columnwidth]{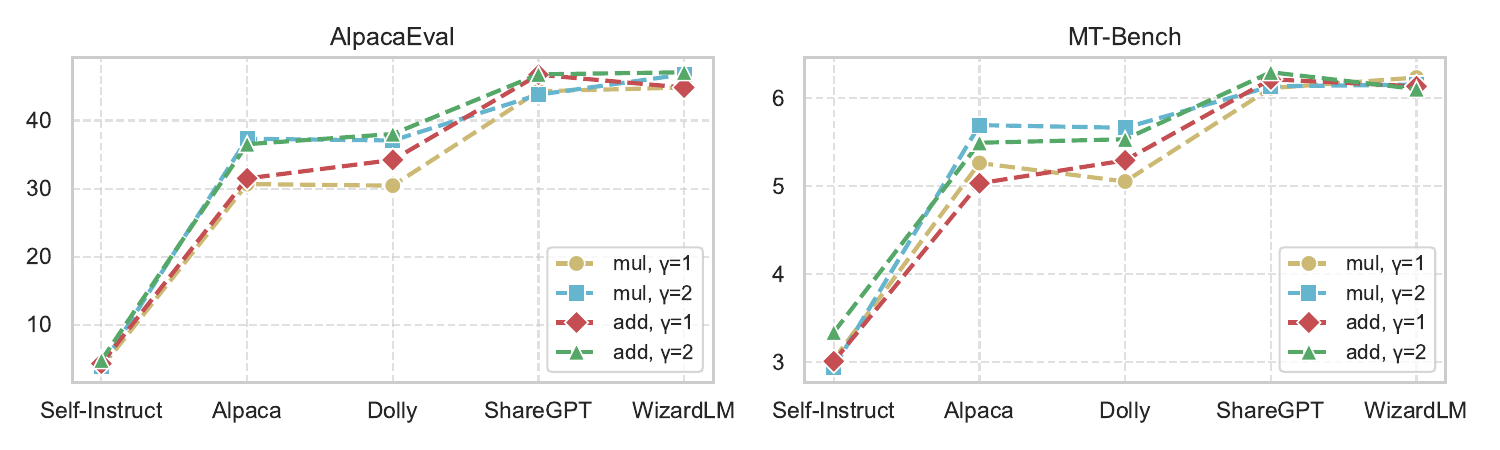}
  \caption{Model performance of different stages during InsBank evolution.}
  \label{fig: evolution}
\end{figure}

In this experiment, we investigate the performance of subsets selected by different data selection methods for model training. Following the temporal order of dataset appearance (i.e. Self-Instruct $\rightarrow$ Alpaca $\rightarrow$ Dolly $\rightarrow$ ShareGPT $\rightarrow$ WizardLM), we performed progressive InsBank evolution using PIBE and take the selected subset for model fine-tuning. The performance of the fine-tuned model across different benchmarks is shown in Figure~\ref{fig: evolution}. 

\subsection{PIBE Hyper-Parameter Analysis}

\begin{figure}[hbtp]
\includegraphics[width=\columnwidth]{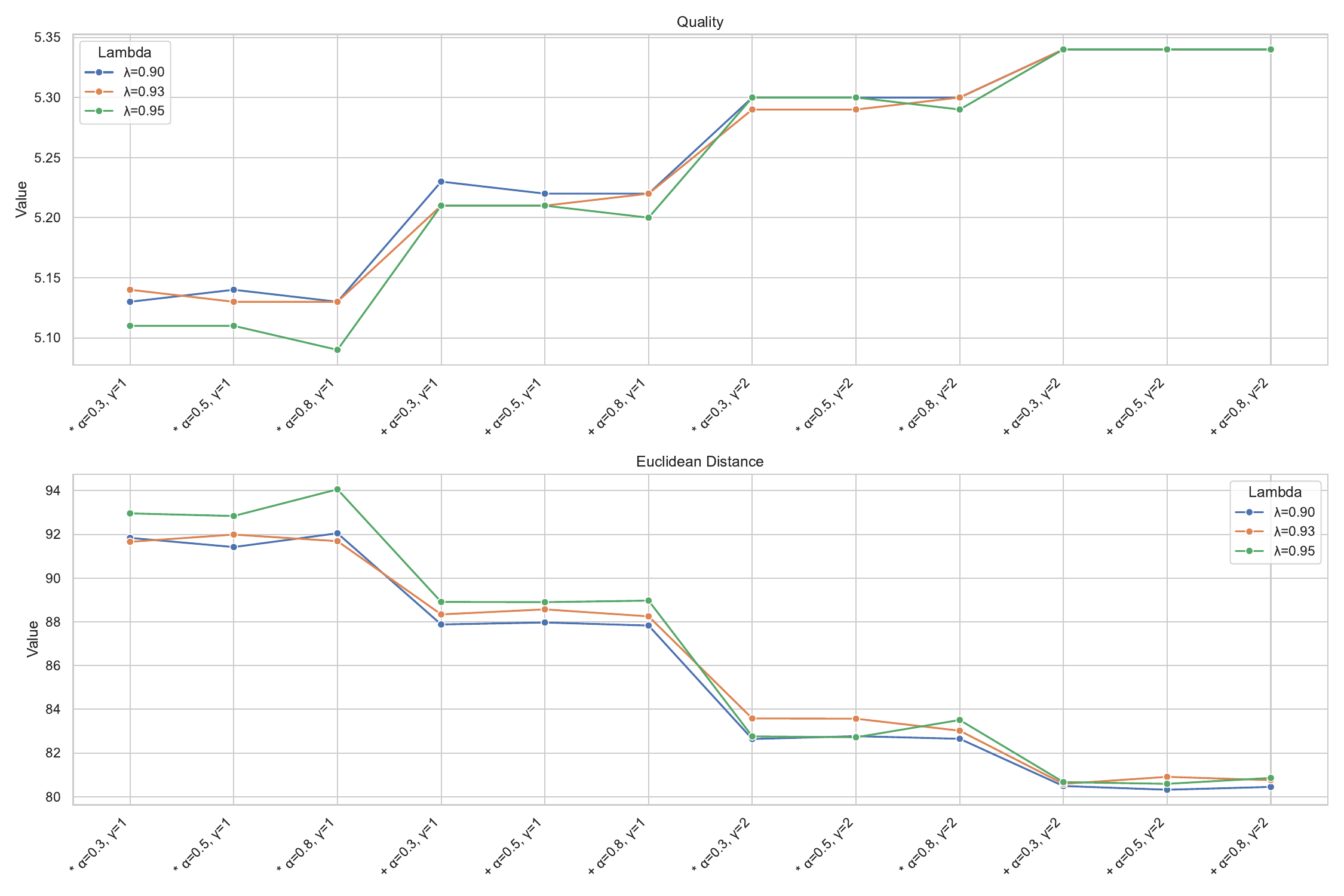}
\caption{InsBank statistics of different hyper-parameters.}
\label{fig: statistics-hyper}
\end{figure}

The damping rate $\beta$ is a hyperparameter inherent to Affinity Propagation, typically set to 0.5, and we have adhered to this default setting. For the analysis of hyperparameters, we focus on examining the quality and diversity of the selected data. We compared different combinations of \(\lambda = [0.9, 0.93, 0.95]\), \(\alpha = [0.3, 0.5, 0.8]\), and \(\gamma = [1, 2]\) in selecting InsBank. The results are shown in Figure~\ref{fig: statistics-hyper}. 
Overall, \(\gamma\) determines the influence of quality on data selection. As \(\gamma\) increases, the average quality of the selected data improves, but diversity decreases. Both \(\lambda\) and \(\alpha\) determine the impact of historical information on the composition of selected data. We find that higher \(\lambda\) and \(\alpha\) values generally result in lower quality but higher diversity in InsBank. This is because, according to the evolution sequence of InsBank, the quality of the data improves progressively. When the influence of historical information increases, more older data is retained in InsBank, leading to relatively lower quality and higher diversity.

We further compare the overlap between the final InsBanks obtained with different hyperparameter. From 0 to 17, the corresponding \([ \alpha, \lambda, \gamma ]\) combinations are as follows: [0.3, 0.90, 1], [0.3, 0.93, 1], [0.3, 0.95, 1], [0.5, 0.90, 1], [0.5, 0.93, 1], [0.5, 0.95, 1], [0.8, 0.90, 1], [0.8, 0.93, 1], [0.8, 0.95, 1], [0.3, 0.90, 2], [0.3, 0.93, 2], [0.3, 0.95, 2], [0.5, 0.90, 2], [0.5, 0.93, 2], [0.5, 0.95, 2], [0.8, 0.90, 2], [0.8, 0.93, 2], [0.8, 0.95, 2]. We observe that when \(\gamma = 2\), the overlap between InsBanks is generally higher compared to when \(\gamma = 1\), due to the increased influence of quality. This observation is reasonable, particularly as \(\gamma\) continues to grows, the results increasingly resemble those of a quality-greedy data selection strategy, where the selection outcomes become fixed regardless of whether historical information is considered. When \(\gamma = 1\), the influence of historical information is relatively more pronounced, resulting in significantly lower overlap rates between different InsBanks compared to when \(\gamma = 2\). Additionally, we observed that when \(\gamma\) and \(\lambda\) are equal, the overlap rates of InsBanks obtained with different \(\alpha\) values are significantly higher than those obtained when \(\gamma\) and \(\alpha\) are equal but with different \(\lambda\) values. This indicates that \(\lambda\) has a greater impact on altering the influence of historical information.

\subsection{Time Costs Analysis}
\label{appendix: time-cost}

We adhered to the data selection settings of the main experiment to compare the actual time costs of data selection between DEITA and PIBE. In this experiment, we ensure that both methods are tested under identical hardware environments. The results are shown in Table~\ref{tab: time-cost-comparison}. It is worth noting that DEITA (full) refers to full-scale data selection, while DEITA (progressive) represents the progressive InsBank Evolution process. Additionally, the time spent loading data is also included in the total time consumption. PIBE achieves higher efficiency compared to DEITA because PIBE's data selection process is parallelized, whereas DEITA requires a sequential traversal of data to perform selection. 

In practice, DEITA's data selection efficiency is primarily influenced by the number of evolution iterations and the size of InsBank. The selection time for DEITA (progressive) grows almost linearly with the number of iterations, while the total data volume has minimal impact. Additionally, as more data is selected into InsBank, the time required to select a new sample increases, as it becomes harder to find a candidate that meets the nearest neighbor similarity constraint. This implies that as the size of InsBank grows, DEITA's efficiency will further decline.

In contrast, PIBE's efficiency is unaffected by the size of InsBank due to its parallelized operations. Instead, the primary factor influencing PIBE's time consumption is the total data volume. An increase in the total data volume leads to a higher number of evolution batches, with each batch requiring approximately 1 minute to process. As a result, PIBE's total data selection time scales linearly with the number of evolution batches.

\begin{table}[htbp]
    \centering
    \small
    \begin{tabular}{lc}
    \toprule
     Method & Time (hrs) \\
     \midrule
     DEITA (full) & 0.68 \\
     DEITA (progressive) & 2.28 \\
     PIBE & 0.21 \\
    \bottomrule
\end{tabular}
\caption{Time costs of DEITA and PIBE.}
\label{tab: time-cost-comparison}
\end{table}

\onecolumn

\newpage
\section{Selected Data Quality Distribution}

\begin{figure*}[h]
\begin{center}
\includegraphics[width=0.98\textwidth]{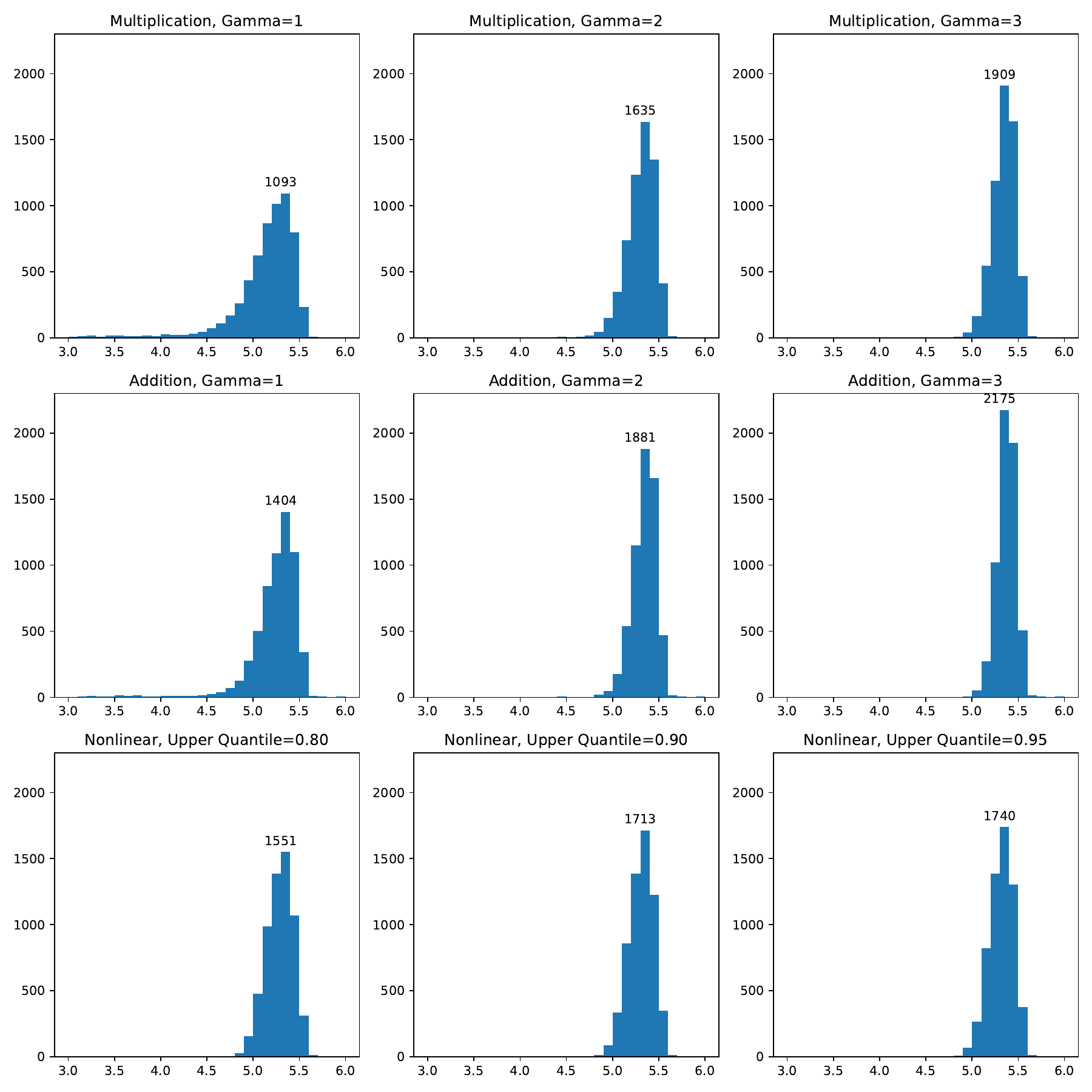}
\end{center}
\caption{Selected data quality distribution of different combination approaches.}
\label{fig: selected-data-quality-distribution}
\end{figure*}

\end{document}